# Algorithm and Hardware Design of Discrete-Time Spiking Neural Networks Based on Back Propagation with Binary Activations

Shihui Yin*, Shreyas K. Venkataramanaiah*, Gregory K. Chen†, Ram Krishnamurthy†, Yu Cao*, Chaitali Chakrabarti*, Jae-sun Seo*

*School of Electrical, Computer and Energy Engineering, Arizona State University, Tempe, AZ, USA
†Circuits Research Lab, Intel Corporation, Hillsboro, OR, USA
Email: syin11@asu.edu

*Abstract*—We present a new back propagation based training algorithm for discrete-time spiking neural networks (SNN). Inspired by recent deep learning algorithms on binarized neural networks, binary activation with a straight-through gradient estimator is used to model the leaky integrate-fire spiking neuron, overcoming the difficulty in training SNNs using back propagation. Two SNN training algorithms are proposed: (1) SNN with discontinuous integration, which is suitable for rate-coded input spikes, and (2) SNN with continuous integration, which is more general and can handle input spikes with temporal information. Neuromorphic hardware designed in 40nm CMOS exploits the spike sparsity and demonstrates high classification accuracy (>98% on MNIST) and low energy (48.4–773 nJ/image).

## I. Introduction

Recently, many deep learning algorithms such as multi-layer perceptrons (MLP) and convolutional neural networks (CNN) have demonstrated human-level recognition accuracy in image and speech classification tasks [1-2]. The number of operations in these artificial neural networks (ANN) do not generally depend on input data, and the high amount of input-independent computations in deep networks can lead to high energy consumption. On the other hand, spiking neural networks (SNN) [3] more closely mimic the operations in biological nervous systems and explore new avenues for brain-like cognitive computing. SNNs can exploit the input-dependent sparsity or redundancy to dynamically scale the amount of computation, leading to energy-efficient hardware implementation [4].

Back-propagation (BP) based stochastic gradient descent is widely used to train ANNs, and has shown high accuracy for many benchmarks. However, literature on SNN training algorithms still has not proven sufficiently high accuracy, especially for deep networks with multiple layers. Existing training algorithms for SNNs are categorized into unsupervised learning (without labeled data) and supervised learning (with labeled training data). Bio-plausible learning rules, such as spike-timing dependent plasticity (STDP) [5-6], have been explored for unsupervised learning, but do not exhibit competitive accuracy for deep networks. Recently, several supervised learning algorithms for SNNs were proposed, where an ANN is first trained using BP and then converted to an SNN [7-9] by mapping real-valued inputs/activations to average firing rates of Poisson spikes. This approach has two drawbacks: (1) it requires many time steps to achieve high accuracy; (2) it is only suitable for rate-coding input spikes, but not for other spike coding formats such as temporal coding. Other supervised learning algorithms have been proposed to directly train SNNs with BP [10-11]. However, these algorithms exhibit complex neuron models with exponential post-synaptic potential or membrane decay, which are computationally expensive.

In this paper, we propose a BP-based training algorithm that can train hardware-friendly SNNs with simple neuron models for both rate-coding and temporal-coding inputs. We also present experimental results of energy-efficient neuromorphic hardware designs in 28nm CMOS that implement our trained SNN models for MNIST [12] and N-MNIST [13] datasets.

## II. Spiking Neuron Models with Binary Activation

In this section, we propose two variants of discrete-time leaky integrate-fire (LIF) spiking neuron models that are suitable for BP-based training of deep SNNs.

### A. Spiking neuron with discontinuous integration (SNN-DC)

In ANNs, the activation output of a neuron $k$ in layer $L$ is:

$$a_k^L = y\left(\sum_i^m \left(a_i^{L-1} w_{i,k}^{L-1}\right) + b_k^L\right), \qquad (1)$$

where $y(x)$ is the activation function, $w_{i,k}^{L-1}$ is the synapse weight connecting neuron $i$ to neuron $k$, and $b_k^L$ is the bias. One popular activation function is rectified linear unit (ReLU), as illustrated in Fig. 1.

Compared to these artificial neurons, spiking neurons have two distinct properties: (1) the neuron value (membrane potential) is integrated over time and (2) each neuron outputs a binary spike. To incorporate these into our first proposed spiking neuron model, we introduce a discrete time step variable $t$ and choose a special binary activation function:

$$v_k^L(t) = \sum_i^m \left(a_i^{L-1}(t) w_{i,k}^{L-1}\right) + b_k^L, \qquad (2)$$

$$a_k^L(t) = y_b\left(v_k^L(t)\right), \qquad (3)$$

where $v_k^L(t)$ is the membrane potential and $y_b(x)$ is a binary activation function depicted in Fig. 1. $y_b(x) = 1$ if $x > \theta$, otherwise $y_b(x) = 0$, where $\theta$ is the firing threshold. Note that this spiking neuron model performs synaptic integration in a *discontinuous* manner, and we denote this model as SNN-DC.

This work was supported in part by the NSF grant 1652866 and Intel Labs.

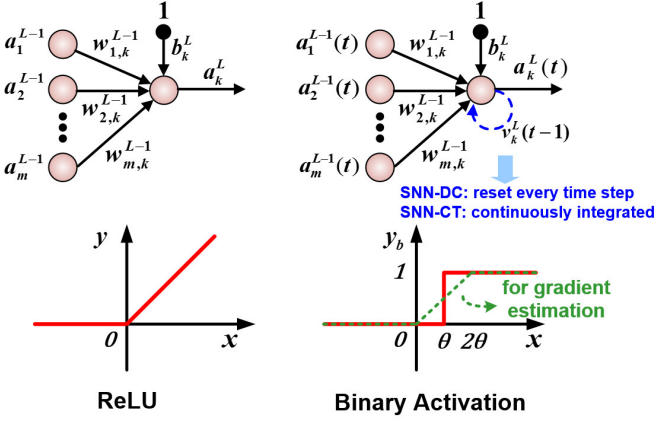

Fig. 1. Comparison of neuron models for ANN with ReLU activation and discrete-time SNN with binary activation (SNN-DC and SNN-CT).

In other words, $v_k^L(t)$ is reset to zero at the beginning of every time step $t$, before the neuron integrates the presynaptic injections and the bias term. If the membrane potential after integration at time step $t$ exceeds threshold $\theta$, the neuron fires ($a_k^L(t) = 1$); otherwise, the neuron remains silent ($a_k^L(t) = 0$).

Since the neuron membrane potential is discontinuous between sequential time steps, time is not incorporated into SNN training. Thus, the SNN-DC model can be reduced to a form similar to (1) except that the activation is a special binary activation function $y_b(x)$. Note that the gradient of this binary function is zero everywhere, blocking back propagation of the error signals. To overcome this difficulty, we adopt the "straight-through estimator" [14] during BP to estimate the gradient of the binary activation function. The straight-through estimator enables fast training with stochastic discrete neurons [14]. When the threshold $\theta = 1$, our straight-through estimator of the gradient of $y_b(x)$ is the following:

$$g_b(x) = \begin{cases} 0.5, & 0 \leq x \leq 2 \\ 0, & otherwise \end{cases} \quad (4)$$

SNNs trained with the SNN-DC model are suitable for rate-coding spike inputs such as Poisson spikes, which assume no temporal correlation between adjacent time steps. Compared to the SNNs converted from ANNs in [7-8], which are also designed for rate-coding spike inputs, our SNNs using the SNN-DC model can achieve good accuracy with much fewer time steps (even with one time step), because the proposed SNNs are directly trained over many single-time-step training samples. Corresponding experimental results will be shown in Section III.

*B. Spiking neuron with continuous integration (SNN-CT)*

For spike input encodings other than rate-coding, the SNN-DC model may not be sufficient to capture the temporal correlation between time steps. By including membrane potential integration across multiple time steps, the SNN-DC model can be extended to a spiking neuron model with *continuous* integration, which we denote as SNN-CT:

$$v_k^L(t^-) = \sum_i^m \left(a_i^{L-1}(t) w_{i,k}^{L-1}\right) + b_k^L + v_k^L(t-1), \quad (5)$$

$$a_k^L(t) = y_b\left(v_k^L(t^-)\right), \quad (6)$$

$$v_k^L(t) = v_k^L(t^-) - \theta \cdot a_k^L(t), \quad (7)$$

where $v_k^L(t^-)$ and $v_k^L(t)$ are the membrane potentials of neuron $k$ at time step $t$ before and after the neuron firing check. The initial membrane potential is set to zero ($v_k^L(-1) = 0$). When $v_k^L(t^-)$ exceeds the threshold $\theta$, the neuron fires and the membrane potential is decremented by $\theta$ as shown in (7) [8]. To enable SNN-CT training with BP-based stochastic gradient descent, the same straight-through estimator shown in (4) is used. Compared to conventional frame-based ANNs, SNN-CT generates outputs with one more dimension: time. We design appropriate loss functions to train SNNs over all the time steps of interest. For example, to train the SNN to generate a specific spike output pattern, the loss is selected as a distance function between the actual and desired output spike patterns. Alternatively, to train rate-encoded outputs over a period of time related to the temporal inputs, the loss is chosen as a function of the sum of output neuron spikes over that period.

## III. SOFTWARE RESULTS ON BENCHMARKS

We trained the discrete-time SNNs with both SNN-DC and SNN-CT neuron models using the Theano framework. In the following experiments, the Adam optimizer [15] is used to train the SNNs, the batch size is 100, and the neuron threshold $\theta = 1$.

*A. SNN-DC evalutaion for MNIST dataset*

The MNIST [12] dataset contains 60k training and 10k testing grey-scale 28×28 pixel images for handwritten digits. The first 50k and last 10k images in the training set are used for training and validation, respectively. We trained both MLP SNNs and convolutional SNNs using SNN-DC neuron models for MNIST. Spike inputs are stochastically sampled from the static image pixels as a Bernoulli process with firing probability proportional to the pixel value normalized to (0, 1). Each training image is presented to the SNN once per epoch. The same training image generates different binary spikes in different epochs as in Bernoulli processes.

The MLP SNNs have two hidden layers with either 256 or 1024 neurons. The squared hinge loss function [16] is employed for training. We performed training for 400 epochs, where learning rate starts at 1e-3 and exponentially decays to 1e-7. Dropout is employed with a dropout ratio of 0.2 for the input layer and 0.1 or 0.3 for the hidden layers of 256 or 1024 neurons, respectively. For baseline comparison, we trained two ReLU-based MLP ANNs of the same network size and then converted them to MLP SNNs following the technique described in [8].

The convolutional SNN architecture is 28×28-12c5-mp2-64c5-mp2-512fc-10o. The two convolution layers have 12 and 64 kernels of size 5×5 (c5), respectively, each followed by a 2×2 max-pooling (mp) layer. After these layers, a fully-connected (fc) layer of 512 neurons is added before the final output layer, and no dropout is employed. Note that max-pooling is often avoided in previous ANN to SNN conversion approaches [7], due to the inefficiency of spike-counting/-coding before/after max-pooling. Our proposed approach does not have this limitation. We trained the convolutional SNN for 200 epochs, with the learning rate decaying from 1e-3 to 1e-5.

Fig. 2 shows MNIST classification accuracies of four MLP SNNs and the convolutional SNN as a function of time steps. The proposed SNNs using SNN-DC achieve high accuracy even with only one time step, whereas the SNNs converted from ANN (ANN2SNN) require 4-16× more time steps to achieve the same accuracy. Our convolutional SNN achieves 99.44% accuracy with 256 time steps and 99.40% accuracy with only 8 time steps.

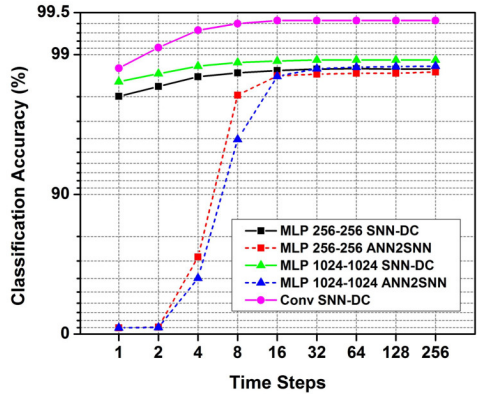

Fig. 2. MNIST accuracies for different time steps (averaged over 20 tests) are shown for various SNN designs.

*B. SNN-CT evaluation for N-MNIST dataset*

The Neuromorphic-MNIST (N-MNIST) [13] dataset is a spiking version of the original frame-based MNIST dataset. N-MNIST dataset images (34×34 pixels) are generated by moving an asynchronous time-based image sensor (ATIS) in front of the MNIST images (Fig. 3). A 3-phase saccadic movement by the ATIS generates two types of events: an on-event for pixel intensity increase; an off-event for decrease.

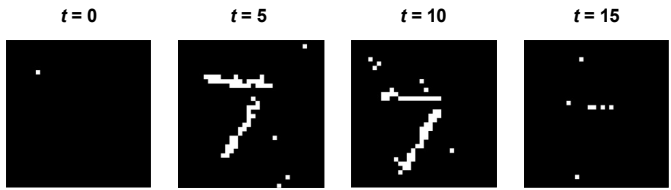

Fig. 3. An example of discrete-time N-MNIST input spikes (digit "7") at four time steps (*t* = 0, 5, 10, 15). Note that the digit "7" is moving downward.

In our experiment, only the on-events in saccade-1 (sensor moving up, digits moving down) are used for training and testing a discrete-time MLP SNN with the SNN-CT model. The 100 ms duration of saccade-1 is equally divided into 16 bins, corresponding to 16 time steps. An input neuron spikes if any on-events of its associated pixel occur in that time bin. To demonstrate the proposed SNN's capability to capture temporal information, we expand the saccade-1 dataset by adding the reversely played data with digits moving upwards. Therefore, training/validation/testing sets now consist of 100k/20k/20k images, with half moving upward and half downward. We trained an MLP SNN consisting of two 256-neuron hidden layers of 256 neurons and a 12-neuron output layer for this augmented N-MNIST benchmark. Note that our MLP SNN is trained for dual tasks with 10 output neurons for digit classification and 2 output neurons for upward and downward motion recognition.

During training, the squared hinge loss function is applied to output spike counts. We trained the SNN for 200 epochs with the learning rate decaying from 1e-3 to 1e-5 and a dropout ratio of 0.2/0.1 for the input/hidden layers. For testing, we classify the digit and its motion direction based on output neuron spike counts. Digit classification and motion recognition accuracies are 96.33% and 99.83%, respectively. This shows that our SNN using the SNN-CT model is successfully trained to extract temporal information from input spikes. For real-time processing with incomplete saccade-1 presentations, the test accuracies after 1 to 16 time steps are shown in Fig. 4. For both tasks, accuracies increase as the number of time steps increase. Fewer input spikes in the last few time steps cause the motion recognition accuracy to slightly decrease.

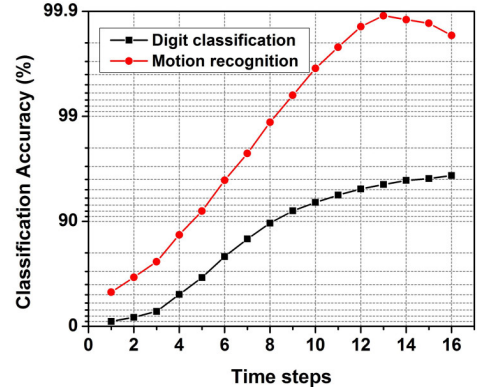

Fig. 4. N-MNIST accuracies of digit classification and motion recognition of the proposed MLP SNN-CT as a function of time steps.

## IV. HARDWARE IMPLEMENTATION RESULTS

We designed neuromorphic hardware for discrete-time MLP SNNs for MNIST and N-MNIST classification tasks in TSMC 28nm HPC CMOS. Both SNNs have two 256-neuron hidden layers. The SNN-DC (SNN-CT) for MNIST (N-MNIST) has 784 (1,156) input neurons and 10 (12) output neurons. Synapse weights are stored in on-chip SRAMs, and the neuron membrane potential values are stored in registers. Fig. 5 shows the overall hardware architecture, where synchronous clocking is used with extensive clock gating. We employ parallel output neurons for the hidden/output layers, while the input spikes of the neurons are processed serially in each clock cycle.

*A. Spike scheduler for event-driven operation*

A spike scheduler featuring 256-/784-/1156-input priority encoders sequentially generates active presynaptic neuron indices from binary input spike vectors. The generated neuron index is sent to the weight memory to fetch the weights for all parallel postsynaptic neurons, which are then integrated onto the neurons' membrane potentials. After all input spikes are processed, the postsynaptic neurons fire if their membrane potentials exceed the threshold. Since only a small fraction (4.8% in the SNN-CT for N-MNIST dataset) of the presynaptic neurons are active at each time step, the spike scheduler enables event-driven computation for only active neuron spikes, substantially reducing the latency and energy.

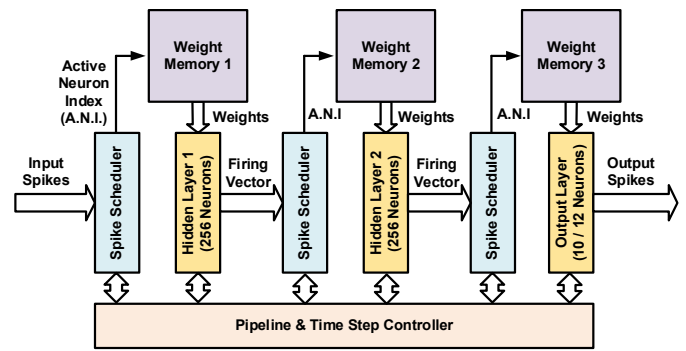

Fig. 5. Hardware achitecture for the proposed MLP SNNs. Total weight memory of SNN-DC (SNN-CT) for MNIST (N-MNIST) is 289 kB (346 kB).

### B. Pipeline architecture and handshaking

All SNN layers are pipelined to improve throughput. Since the number of active spikes vary with layer and time, handshake signals are exchanged between adjacent layers to reduce the overall latency. Each layer starts processing the its input spikes for next time step when it receives a 'done' signal from its previous layer and a 'data_fetched' signal from its next layer.

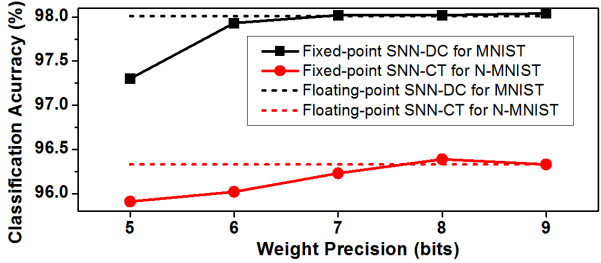

Fig. 6. Classification accuracies for different precision of weights.

### C. Precision, area, latency, and energy

Fig. 6 shows how classification accuracy varies with weight precision. We selected 7-b weight precision for both SNN-DC and SNN-CT, resulting in negligible accuracy loss compared to floating-point precision. Synapse weights are stored in SRAM generated from a memory compiler, and digital logic is synthesized using standard cells. The total post-layout neuromorphic processor area is 1.65 mm$^2$, with 0.79 mm$^2$ logic and 0.86 mm$^2$ memory. At the nominal supply voltage of 0.9 V, power consumption results are obtained from Cadence Innovus with data switching activity information from post-layout simulation. The proposed SNN hardware implementation results are summarized in Table I, where the capability to train/classify with a small number of time steps greatly reduced the energy down to 48.4 nJ per classification.

Fig. 7 shows a comparison to previous MNIST hardware designs [17-19] for accuracy and energy. Compared to a recent 28nm ANN design [17], the proposed SNN-DC reduces energy by ~3X at iso-accuracy between 98% and 99%. Note that our reported energy is based on post-layout simulation, while others are based on chip measurement results.

TABLE I. SNN HARDWARE IMPLEMENTATION RESULTS

| SNN Design | # of time steps | Freq. (MHz) | Latency (cycles) | Power (mW) | Energy (nJ) per classification |
|---|---|---|---|---|---|
| SNN-DC for MNIST | 1 | 163 | 112 | 70.4 | 48.4 |
| | 16 | 163 | 1780 | 70.8 | 773 |
| SNN-CT for N-MNIST | 16 | 163 | 654 | 73.2 | 294 |

## V. CONCLUSION

In this paper, we presented new training algorithms and classification hardware design for discrete-time SNNs. By employing binary activation with a straight-through estimator, we can effectively train SNNs using error back propagation. SNN-DC with discontinuous neuron integration is suitable for rate-coded input spikes, and SNN-CT with continuous neuron integration is more general and applicable for input spikes with temporal information. The corresponding MLP SNNs were implemented in 28nm CMOS, demonstrating high accuracy and low energy. SNN-DC shows 98.0–98.70% accuracy at 48.4–773 nJ per classification for MNIST, and SNN-CT shows 96.33% accuracy at 294 nJ per classification for N-MNIST.

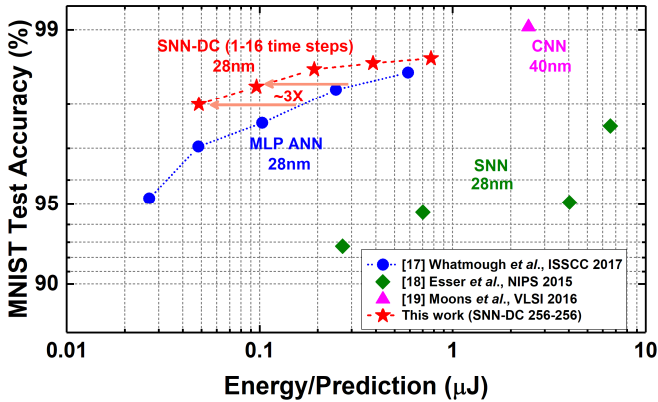

Fig. 7. MNIST accuracy and energy comparison to hardware design literature.